%% file: gfl.tex
\documentclass[11pt]{article}%
\usepackage{amsfonts}
\usepackage{amssymb}
\usepackage{graphicx}
\usepackage{caption}
\usepackage{subcaption}
\usepackage{setspace}
\usepackage[square]{natbib}
\usepackage{amsmath}%
\usepackage{amsthm}%
\usepackage{url}
\usepackage{palatino}
\usepackage{paralist}
\usepackage[top=8pc,bottom=8pc,left=8pc,right=8pc]{geometry}
\usepackage{dblfloatfix}
\usepackage{algorithm}
\usepackage[noend]{algpseudocode}

\newtheorem{theorem}{Theorem}

\newtheorem{definition}{Definition}

\newcommand{\R}{\mathcal{R}}

\newcommand{\vnorm}[1]{\left|\left|#1\right|\right|}

\def\grad{\nabla}

\title{A Fast and Flexible Algorithm for the Graph-Fused Lasso}
\author{
Wesley Tansey\footnote{Department of Computer Science, University of Texas at Austin, \texttt{tansey@cs.utexas.edu} (corresponding author)} \\
James G.~Scott\footnote{Department of Information, Risk, and Operations Management; Department of Statistics and Data Sciences; University of Texas at Austin, \texttt{james.scott@mccombs.utexas.edu}}
}
\date{This version: \today}

\begin{document}

\maketitle

\begin{abstract} 
We propose a new algorithm for solving the graph-fused lasso (GFL), a method for parameter estimation that operates under the assumption that the signal tends to be locally constant over a predefined graph structure. Our key insight is to decompose the graph into a set of trails which can then each be solved efficiently using techniques for the ordinary (1D) fused lasso. We leverage these trails in a proximal algorithm that alternates between closed form primal updates and fast dual trail updates. The resulting techinque is both faster than previous GFL methods and more flexible in the choice of loss function and graph structure. Furthermore, we present two algorithms for constructing trail sets and show empirically that they offer a tradeoff between preprocessing time and convergence rate.
\end{abstract} 

\section{Introduction}
\label{sec:introduction}

\input{introduction}

\section{ADMM via Graph Decomposition}
\label{sec:decomposition}

\input{decomposition}

\subsection{Optimization}
\label{sec:optimization}

\input{optimization}

\section{Trail decompositions}
\label{sec:grid_example}

\input{grid_example}

\section{Experiments}
\label{sec:experiments}

\input{experiments}

\section{Discussion}
\label{sec:discussion}

\input{discussion}

\section{Acknowledgments}
\label{sec:acknowledgements}

We thank Ryan Tibshirani for the initial idea of decomposing a grid graph by its rows and columns and solving via a proximal algorithm; the generalization of that idea to generic graphs led to this paper.

\appendix

\section{Gradient Smoothing Alternative}
\label{sec:gradientsmoothing}

\input{spg}

\begin{small}
\singlespacing
\bibliographystyle{abbrvnat}
\bibliography{gfl}
\end{small}

\end{document}

%% file: introduction.tex
Let $\boldsymbol\beta = \{\beta_i : i \in \mathcal{V}\}$ be a signal whose components are associated with the vertices of an undirected graph $\mathcal{G} = (\mathcal{V}, \mathcal{E})$.  Our goal is to estimate $\boldsymbol\beta$ on the basis of noisy observations $\mathbf{y}$.  This paper proposes a new algorithm for solving the graph-fused lasso (GFL), a method for estimating $\boldsymbol\beta$ that operates under the assumption that the signal tends to be locally constant over $\mathcal{G}$.  The GFL is defined by a convex optimization problem that penalizes the first differences of the signal across edges:
\begin{equation}
\label{eqn:gfl_loss_function}
\begin{aligned}
& \underset{\boldsymbol\beta \in \R^n}{\text{minimize}}
& & 
\ell(\mathbf{y}, \boldsymbol\beta) + \lambda \sum_{(r,s) \in \mathcal{E}} |\beta_r - \beta_s|  \, ,
\end{aligned}
\end{equation}
where $\ell$ is a smooth convex loss function and $n = |\mathcal{V}|$.

Many special cases of this problem have been studied in the literature.  For example, when $\ell$ is squared error loss and $\mathcal{G}$ is a one-dimensional chain graph, (\ref{eqn:gfl_loss_function}) is the ordinary (1D) fused lasso \citep{tibs:fusedlasso:2005}. When $\mathcal{G}$ is a two- or three-dimensional grid graph, (\ref{eqn:gfl_loss_function}) is often referred to as total-variation denoising \citep{rudin:osher:faterni:1992}.  In both cases, there exist very efficient algorithms for solving the problem.  In the 1D case (i.e.~where $\mathcal{E}$ defines a chain graph), a dynamic programming routine  due to \citet{johnson:2013} can recover the exact solution to \eqref{eqn:gfl_loss_function} in linear time.  And when the graph corresponds to a 2D or 3D grid structure with distance-based edge weights and a Gaussian loss, a parametric max-flow algorithm \citep{chambolle:etal:2009} can be used to rapidly solve \eqref{eqn:gfl_loss_function} to a desired precision level. Furthermore, when the loss is Gaussian but no grid structure is present in the graph, the GFL is a specialized case of the generalized lasso \citep{tibs:taylor:2011}.

However, none of these approaches apply to the case of an arbitrary smooth convex loss function and a generic graph (which need not be linear or even have a geometric interpretation).  In this paper, we propose a method for solving \eqref{eqn:gfl_loss_function} in this much more general case.   The key idea is to exploit a basic result in graph theory that allows any graph to be decomposed into a set of trails. After forming such a decomposition, we solve (\ref{eqn:gfl_loss_function}) using a proximal algorithm that alternates between a closed-form primal update and a dual update that involves a set of independent 1-dimensional fused-lasso subproblems, for which we can leverage the linear-time solver of \citet{johnson:2013}.   The resulting algorithm is applicable to generic graphs, flexible in the choice of loss function $\ell$, and fast.  To our knowledge, no existing algorithms for the graph-fused lasso satisfy all three criteria.  A relevant analogy is with the paper by \cite{ramdas:tibs:2014}, who also exploit this 1D solver to derive a fast ADMM for trend filtering and which can be solved on graphs \citep{wang:etal:2015}. 

We present two variations on the overall approach (corresponding to two different algorithms for decomposing graphs into trails), and we investigate their performance empirically.  Our results reveal that these two variations offer a choice between quickly generating a minimal set of trails at the expense of longer convergence times, versus spending more time on careful trail selection in order to achieve more rapid convergence. We further benchmark our approach against alternative approaches to GFL minimization \citep{wahlberg:etal:2012,chambolle:etal:2009,tansey:etal:2014} and show that our trail-based approach matches or outperforms the state of the art on a collection of real and synthetic graphs.  For the purposes of exposition, we use the typical squared loss function $\ell(\mathbf{y}, \boldsymbol\beta) = \frac{1}{2} \sum_i (y_i - \beta_i)^2$, though our methods extend trivially to other losses involving, for example, Poisson or binomial sampling models.


The remainder of the paper is organized as follows. Section \ref{sec:decomposition} presents our graph decomposition approach and derives an ADMM optimization algorithm.  Section \ref{sec:grid_example} details our two algorithms for discovering trails and demonstrates the effectiveness of good trails with a simple illustrative example.  Section \ref{sec:experiments} presents our benchmark experiments across several graph types. Finally, Section \ref{sec:discussion} presents concluding remarks.

%% file: decomposition.tex
\subsection{Background}
\label{sec:background}
The core idea of our algorithm is to decompose a graph into a set of trails. We therefore begin by reviewing some basic definitions and results on trails in graph theory. Let $\mathcal{G} = (\mathcal{V}, \mathcal{E})$ be a graph with vertices $\mathcal{V}$ and edge set $\mathcal{E}$.  We note two preliminaries.  First, every graph has an even number of odd-degree vertices \citep{west:2001}.  Second, if $\mathcal{G}$ is not connected, then the objective function is separable across the connected components of $\mathcal{G}$, each of which can be solved independently. Therefore, for the rest of the paper we assume that the edge set $\mathcal{E}$ forms a connected graph.

\begin{definition}
A \textbf{walk} is a sequence of vertices, where there exists an edge between the preceding and following vertices in the sequence.
\end{definition}

\begin{definition}
A \textbf{trail} is a walk in which all the edges are distinct.
\end{definition}

\begin{definition}
An \textbf{Eulerian trail} is a trail which visits every edge in a graph exactly once.
\end{definition}

\begin{definition}
A \textbf{tour} (also called a \textbf{circuit}) is a trail that begins and ends at the same vertex.
\end{definition}

\begin{definition}
An \textbf{Eulerian tour} is a circuit where all edges in the graph are visited exactly once.
\end{definition}

\begin{theorem}[\cite{west:2001}, Thm 1.2.33]
\label{thm:num_trails}
The edges of a connected graph with exactly $2k$ odd-degree vertices can be partitioned into $k$ trails if $k > 0$. If $k = 0$, there is an Eulerian tour. Furthermore, the minimum number of trails that can partition the graph is \texttt{max}$(1, k)$.
\end{theorem}

Theorem \ref{thm:num_trails} reassures us that any connected graph can be decomposed into a set of trails $\mathcal{T}$ on which our optimization algorithm can operate.  Specifically, it allows us to re-write the penalty function as
$$
\sum_{(r,s) \in \mathcal{E}} |\beta_r - \beta_s|  = \sum_{t \in \mathcal{T}} \sum_{(r,s) \in t} |\beta_r - \beta_s| \, .
$$
We first decribe the algorithmic consequences of this representation, and then we turn to the problem of how to actually perform the decomposition of $\mathcal{G}$ into trails.

%% file: optimization.tex
Given a graph decomposed into a set of trails, $\mathcal{T} = \{ t_1, t_2, \ldots, t_k \}$, we can rewrite \eqref{eqn:gfl_loss_function} by grouping each trail together,

\begin{equation}
\label{eqn:trail_loss_function}
\begin{aligned}
& \underset{\boldsymbol\beta \in \R^n}{\text{minimize}}
& & 
\ell(\mathbf{y}, \boldsymbol\beta) + \lambda \sum_{t \in \mathcal{T}} \sum_{(r,s) \in t} |\beta_r - \beta_s|  \, .
\end{aligned}
\end{equation}

For each trail $t$ (where $|t| = m$), we introduce $m+1$ slack variables, one for each vertex along the trail. If a vertex is visited more than once in a trail then we introduce multiple slack variables for that vertex.  We then re-write (\ref{eqn:trail_loss_function}) as
\begin{equation}
\label{eqn:trail_loss_function_duals}
\begin{aligned}
& \underset{\boldsymbol\beta \in \R^n}{\text{minimize}}
& & 
\ell(\mathbf{y}, \boldsymbol\beta) + \lambda \sum_{t \in \mathcal{T}} \sum_{(r,s) \in t} |z_r - z_s|\, . \\
& \text{subject to}
& & \beta_r = z_r \\
& & & \beta_s = z_s \, .
\end{aligned}
\end{equation}
We can then solve this problem efficiently via an ADMM routine \citep{boyd:etal:2011} with the following updates:

\begin{align}
\label{eqn:admm_updates_beta}
\boldsymbol\beta^{k+1} & = \underset{\boldsymbol\beta}{\text{argmin}} \left( \ell(\mathbf{y}, \boldsymbol\beta) + \frac{\alpha}{2}\vnorm{A\boldsymbol\beta - \mathbf{z}^k + \mathbf{u}^k}^2 \right) \\
\label{eqn:admm_updates_z}
\mathbf{z}^{k+1}_t & = \underset{\mathbf{z}}{\text{argmin}} \left( w \sum_{r \in t} (\tilde{y}_r - z_r)^2 + \sum_{(r,s) \in t} |z_r - z_s|  \right) \; , \quad t \in \mathcal{T} \\
\label{eqn:admm_updates_u}
\mathbf{u}^{k+1}& = \mathbf{u}^k + A\boldsymbol\beta^{k+1} - \mathbf{z}^{k+1}
\end{align}
where $u$ is the scaled dual variable, $\alpha$ is the scalar penalty parameter, $w = \frac{\alpha}{2}$, $\tilde{y}_r = \beta_r - u_r$, and $A$ is a sparse binary matrix used to encode the appropriate $\beta_i$ for each $z_j$.  Here we use $t$ to denote both the vertices and edges along trail $t$, so that the first summation in (\ref{eqn:admm_updates_z}) is over the vertices while the second summation is over the edges.

In the concrete case of squared-error loss, $\ell(\mathbf{y}, \boldsymbol\beta) = \sum_i (y_i - \beta_i)^2$, the $\boldsymbol\beta$ updates have a simple closed-form solution:
\begin{equation}
\label{eqn:admm_updates_beta_squared_loss}
\beta_i^{k+1} = \frac{2y_i + \alpha \sum_{j \in \mathcal{J}} (z_j - u_j)}{2 + \alpha |\mathcal{J}|} \, ,
\end{equation}
where $\mathcal{J}$ is the set of dual variable indices that map to $\beta_j$. Crucially, our trail decomposition approach means that each trail's $z$ update in \eqref{eqn:admm_updates_z} is a one-dimensional fused lasso problem which can be solved in linear time via an efficient dynamic programming routine \citep{johnson:2013}.

The combination of the closed-form updates in the $\boldsymbol\beta$- and $\mathbf{u}$-steps, and the linear-time updates in the $z$-step, make the overall algorithm highly efficient in each of its components. The dominating concern therefore becomes the number of alternating steps required to reach convergence, which is primarily influenced by the choice of graph decomposition.  We turn to this question in Section \ref{sec:grid_example}, before describing the results of several experiments designed to evaluate the efficiency of our ADMM algorithm across a variety of graph types and trail decomposition strategies.



%% file: grid_example.tex

\subsection{Choosing trails: two approaches}
\label{sec:decomposition_algorithms}
Two approaches to proving the first half of Theorem \ref{thm:num_trails} are given in \citep{west:2001}, and each corresponds to a variation of our algorithm. The first approach is to create $k$ {``pseudo-edges''} connecting the $2k$ odd-degree vertices, and then to find an Eulerian tour on the surgically altered graph.  To decompose the graph into trails, we then walk along the tour (which by construction enumerates every edge in the original graph exactly once). Every time a pseudo-edge is encountered, we mark the start of a new trail. See Algorithm \ref{alg:pseudo_tour} for a complete description of the approach. The second approach is to iteratively choose a pair of odd-degree vertices and remove a shortest path connecting them. Any component that is disconnected from the graph then has an Eulerian tour and can be appended onto the trail at the point of disconnection.

\begin{algorithm}[t]
\caption{Pseudo-tour graph decomposition algorithm. Yields a minimal set of trails but does not control trail length variance.}
\label{alg:pseudo_tour}
\begin{algorithmic}[1]
    \Require Graph $\mathcal{G} = (\mathcal{V}, \mathcal{E})$
    \Ensure Set of trails $\mathcal{T}$
    \State Pseudo-edge set $\tilde{\Gamma} \gets \emptyset$
    \State $\mathcal{\tilde{G}} \gets \mathcal{G}$
    \State $\tilde{V} \gets \text{OddDegreeVertices}(\mathcal{G})$
    \While{$\tilde{V} > 0$}
        \State Choose two unadjacent vertices, $\nu_i, \nu_j \in \tilde{V}$
        \State Remove $\nu_i, \nu_j$ from $\tilde{V}$
        \State Add pseudo-edge $\tilde{e} = (\nu_i, \nu_j)$ to $\mathcal{\tilde{G}}$ and $\tilde{\Gamma}$
    \EndWhile
    \State Find a tour over the pseudo-graph, $\tilde{t} \gets EulerianCircuit(\mathcal{\tilde{G}})$
    \State Create $\mathcal{T} = \{t_1, t_2, \ldots, t_k\}$ by breaking $\tilde{t}$ at every edge in $\tilde{\Gamma}$
    \State
    \Return $\mathcal{T}$
\end{algorithmic}
\end{algorithm}

Both of these approaches are generic graph decomposition algorithms and are not tailored for our specific application. Algorithm \ref{alg:pseudo_tour} is appealing from a computational perspective, as finding an Eulerian tour can be done in linear time \citep{fleischner:etal:1990}. The latter approach is computationally more intensive during the decomposition phase, but enables control over how to choose each odd-degree pair and thus allows us greater control over the construction of our trails. In the case where the graph structure remains fixed across multiple problem instances, it may be worthwhile to spend a little extra time in the decomposition phase if it yields trails that will reduce the average or worst case time to convergence. This leaves an open question: \textbf{what makes a good trail set?}

Given that our dual updates are quick and exact, one intuitive goal is to find well-balanced trail sets where all of the trails are fairly long and can maximally leverage the linear-time solver. Finding the trail decomposition that maximizes this metric is in general a non-trivial problem, since removing an edge will alter the structure of the subgraph and may change the structure of the trails added in following iterations. Moreover, exhaustively exploring the effects of adding each edge is unlikely to be tractable even for small graphs. We therefore choose a heuristic that balances greedy trail selection with potential downstream effects. For instance, adding the maximum-length trail at each iteration would add a single long trail to the resulting trail set, but could potentially leave many short trails remaining. Conversely, adding the minimum-length trail is clearly not a good choice as it will generate many short trails directly. We found that choosing the median-length trail tends to perform well empirically across a range of graph types, and this is the heuristic used in Algorithm \ref{alg:median_trails}.\footnote{For computational purposes, Algorithm \ref{alg:median_trails} handles the graph split merging edge-case by simply creating two separate trails. The result is a small, but not necessarily minimal, set of trails, unlike Algorithm \ref{alg:pseudo_tour}, which yields a provably minimal set of trails.}

\begin{algorithm}[t]
\caption{Median trails heuristic algorithm. Tries to find a set of trails with a high mean-to-variance length ratio.}
\label{alg:median_trails}
\begin{algorithmic}[1]
    \Require Graph $\mathcal{G} = (\mathcal{V}, \mathcal{E})$
    \Ensure Set of trails $\mathcal{T}$
    \State Subgraphs set $S \gets \{ \mathcal{G} \}$
    \State $\mathcal{T} \gets \emptyset$
    \While{$|S| > 0$}
        \ForAll{subgraph $s = (\nu, \epsilon)$ in $S$}
            \State $\mathbf{\tilde{\nu}} \gets \text{OddDegreeVertices}(s)$
            \If{$|\mathbf{\tilde{\nu}}| = 0$}
                \State $t \gets \text{EulerianCircuit}(s)$
            \ElsIf{$|\mathbf{\tilde{\nu}}| = 2$}
                \State $t \gets \text{EulerianTour}(\mathbf{\tilde{\nu}}_1, \mathbf{\tilde{\nu}}_2)$
            \Else
                \State $\mathcal{P} \gets \emptyset$
                \ForAll{pair ($\nu_i, \nu_j$) \textbf{in} $\text{PairCombinations}(\mathbf{\tilde{\nu}})$}
                    \State $\mathcal{P} \gets \mathcal{P} \cup \text{ShortestPath}(\epsilon, \nu_i, \nu_j)$
                \EndFor
                \State $t \gets \text{MedianLengthPath}(\mathcal{P})$
            \EndIf
            \State $\mathcal{T} \gets \mathcal{T} \cup t$
            \State $\text{CheckSubgraph}(S, s)$
        \EndFor
    \EndWhile
    \State
    \Return $\mathcal{T}$
\end{algorithmic}
\end{algorithm}

\subsection{Grid-graph example}

It may not be immediately clear that the trail decomposition strategy makes any difference to the optimization algorithm. After all, both strategies produce trails consisting of the same number of total edges. Even in extreme cases when it seems obvious that it would likely have an effect (e.g. a single long trail vs. a set of single edges), it is hard to know the extent to which the graph decomposition can impact algorithm performance. To help build this intuition, we delay the description of our optimization algorithm and instead first present a motivating example.

\begin{figure}[tbh]
\begin{subfigure}{.65\textwidth}
  \centering
  \includegraphics[width=\textwidth]{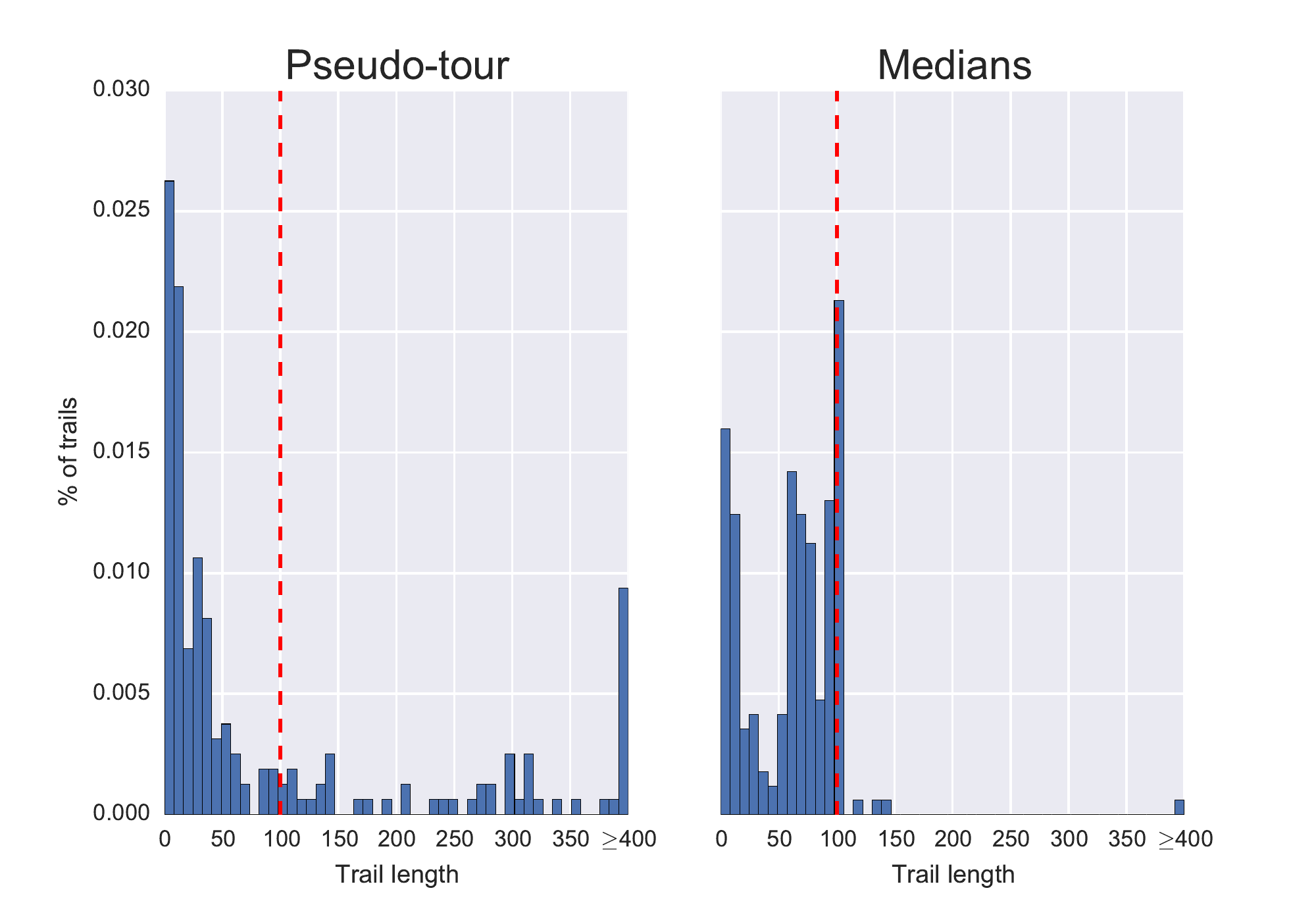}
  \caption{}
  \label{fig:grid_example_distributions}
\end{subfigure}%
\begin{subfigure}{.1\textwidth}
  \centering
  \begin{tabular}{|lr|} \hline
  Trails & Steps \\ \hline
  Rows+Cols & 407.100 \\ & (+/- 13.079) \\ \hline
  Pseudo-tour & 1506.820 \\ & (+/- 42.261) \\ \hline
  Medians & 1073.690 \\ & (+/- 31.062) \\ \hline
  \end{tabular}
  \caption{}
  \label{fig:grid_example_performance}
\end{subfigure}
\caption{(Left) The distribution of trail sizes for the two graph decomposition algorithms on a $100\times100$ grid graph. The Pseudo-tour algorithm produces a large number of short and long trails, where as the Medians algorithm generates more moderately sized trails. The red line marks the size of every trail if using a simple row and column trail set (high mean:variance ratio). (Right) The mean number of steps to convergence across 100 independent trials on the $100\times100$ grid graph, with standard error in parentheses.}
\label{fig:grid_example}
\end{figure}

Consider a simple $100\times100$ grid graph example, where each node has an edge to its immediate neighbors. Figure \ref{fig:grid_example_distributions} shows the distribution of trail lengths for the results of both Algorithms \ref{alg:pseudo_tour} (Pseudo-tour) and \ref{alg:median_trails} (Medians). Clearly the pseudo-tour trails are much more dispersed, producing a large number of short ($< 50$ nodes) and long ($> 150$ nodes) trails. Conversely, the medians algorithm generates mostly moderate trails of length approximately 100. For a grid graph, it is also straightforward to hand-engineer a very well-balanced set of trails: simply use each of the 100 rows and 100 columns as trails. This produces a small set of trails each of length 100; the row and column trails are noted via the dashed red line in the figure.

The results of 100 independent trials using each of the trail sets are displayed in the table in Figure \ref{fig:grid_example_performance}. The difference in performance is striking: the row and column trails converge nearly four times faster than the pseudo-tour trails.\footnote{In general, step counts are directly translatable into wall-clock time for a serial implementation of our algorithm. We choose to show steps when comparing trails as they are not effected by externalities like context switching, making them slightly more reliable.} The medians decomposition also performs substantially worse than the row and column trails, but nearly 1/3 better than the pseudo-tour trails. Clearly, the trail choice can have a dramatic impact on convergence time, especially for large-scale problems where a 30-60\% speedup could be a substantial improvement.

Despite the performance differences, it is important to remember that the two decomposition algorithms have different goals. The pseudo-tour approach of Algorithm \ref{alg:pseudo_tour} aims to quickly find a minimal set of trails, which maximizes the mean trail length. The downside is the risk of high variance in the trail lengths and potentially slower convergence. Conversely, the medians heuristic in Algorithm \ref{alg:median_trails} attempts to select trails that will be well-balanced, but comes with a high preprocessing cost since it must consider all $2(k-t)\times2(k - t- 1)$ paths at every $t^{\text{th}}$ step and requires $k$ steps. While in practice we can randomly sample a subset of candidate pairs for large graphs, the difference in preprocessing times between the two algorithms can still be massive: for the DNA example in Section \ref{sec:experiments}, Algorithm \ref{alg:median_trails} takes hours while Algorithm \ref{alg:pseudo_tour} finishes in less than ten seconds.



%% file: experiments.tex
\subsection{Alternative approaches}
\label{sec:alternative_approaches}
To evaluate the effectiveness of our algorithm, we compare it against several other approaches to solving \eqref{eqn:gfl_loss_function}.\footnote{In our effort to be thorough, we also attempted to evaluate against smoothing proximal gradient \citep{chen:etal:2012}, but were unable to achieve adequate convergence with this method (see Appendix \ref{sec:gradientsmoothing} for details).}

In one method used in the recent literature \citep{wahlberg:etal:2012, tansey:etal:2014}, two sets of slack variables are introduced in order to reformulate the problem such that the dominant cost is in repeatedly solving the system $(I + L)x = b$, where $L$ is the graph Laplacian and only $b$ changes at every iteration. Modern algebraic multigrid (AMG) methods have become highly efficient at solving such systems for certain forms of $L$, and in the following benchmarks we compare against an implementation of this approach using a smoothed aggregation AMG approach \citep{pyamg:2014} with conjugate gradient acceleration. In all reported timings, we do not include the time to calculate the multigrid hierarchy, though we note this can add significant overhead and may even take longer than solving the subsequent linear system.

In the spatial case, such as the grid graph from Section \ref{sec:grid_example}, a highly efficient parametric max-flow method is available \citep{chambolle:etal:2009}. It is also straight-forward in many spatial cases to hand-engineer trails, as in our row and column trails from the grid graph example. Indeed, the row and column trails are equivalent to the ``proximal stacking'' method of \citep{barbero:sra:2014}, which uses a proximal algorithm with the row and columns split and updated with an efficient 1D solver for anisotropic total variation denoising on geometric graphs. In our benchmarks, we consider both the max-flow and row and column trails, to give an idea of performance improvement when special structure can be leveraged in the graph.

We also test different trail decomposition strategies. To demonstrate the benefits of our trail decomposition strategy, we compare against a naive decomposition that simple treats each edge as a separate trail. We also demonstrate the advantage of our median heuristic by testing it against a similar heuristic that randomly adds trails. Finally, we show how performance of the row and column trails degrades as their length decreases, to illustrate the benefit of longer trails. For all trail decomposition strategies, we solve the 1D fused lasso updates using the dynamic programming solution of \citep{johnson:2013} implemented in the \texttt{glmgen} package.\footnote{\url{https://github.com/statsmaths/glmgen}}

\subsection{Performance criteria}
\label{sec:performance_criteria}
For all experiments, we report timings averaged across 30 independent trials; for trail decomposition strategy comparisons, we report the number of steps rather than wall-clock time. In each trial, we randomly create four blobs of different mean values, with each blob being approximately $5\%$ of the nodes in the graph. The algorithms are then run until $10^-4$ precision and we use a varying penalty acceleration heuristic \citep{boyd:etal:2011}, though performance results are not meaningfully different for different precision levels or fixed penalty parameters.

\begin{figure}[tbh]
\centering
    \begin{subfigure}{.45\textwidth}
      \centering
      \frame{\includegraphics[width=\textwidth]{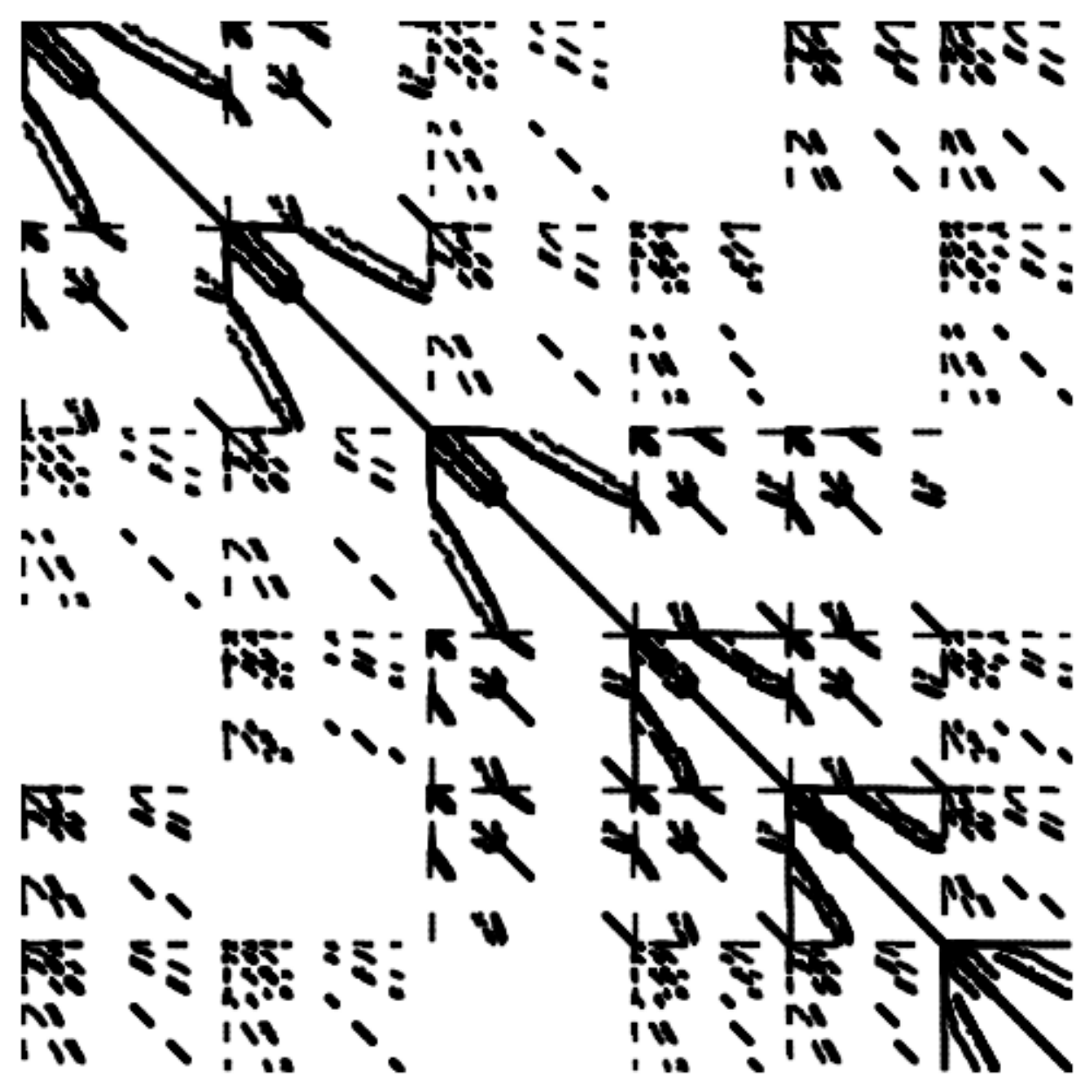}}
      \caption{Vibrobox}
      \label{fig:vibrobox_connectivity}
    \end{subfigure}
    \begin{subfigure}{0.1\textwidth}
    \end{subfigure}
    \begin{subfigure}{.45\textwidth}
      \centering
      \frame{\includegraphics[width=\textwidth]{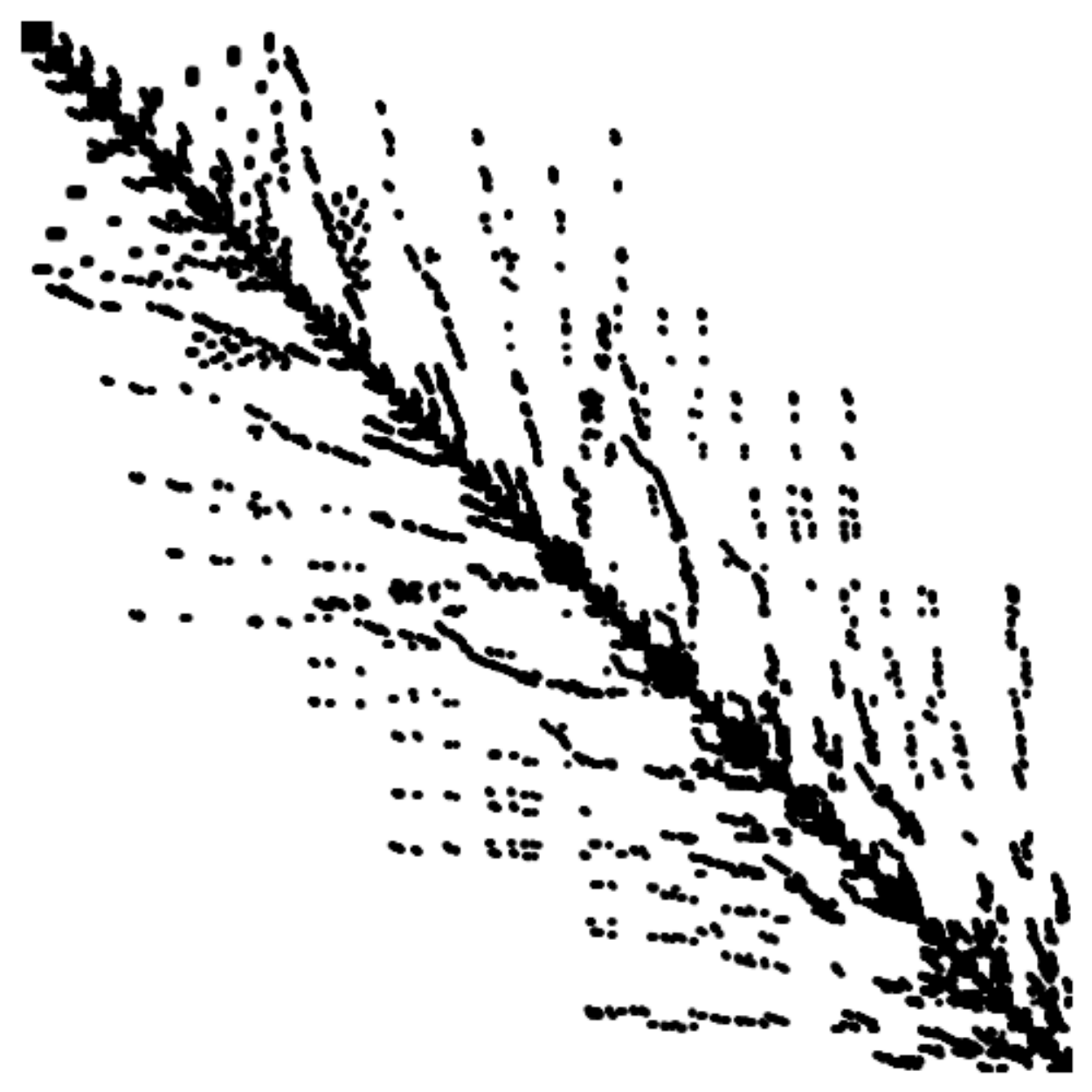}}
      \caption{DNA Electrophoresis}
      \label{fig:dna_connectivity}
    \end{subfigure}%
\caption{The connectivity structure of the two real-world graphs; black cells denote connected nodes. \ref{fig:vibrobox_connectivity}) An accoustic vibrobox simulation with 12K nodes. \ref{fig:dna_connectivity}) A DNA electrophoresis model with 39K nodes.} 
\label{fig:real_graphs}
\end{figure}

With the exception of the spatially-structured methods, all algorithms are compared on two synthetic and two ``real-world'' graphs. For the synthetic examples, we create randomly-connected graphs with approximately $99.8\%$ sparsity, and square grids with adjacency defined by each node's immediate neighbor; performance is then tracked as the number of nodes in the graphs increases. However, randomly generated networks often produce graph forms that are unlikely to be seen in the wild; conversely, square grids are likely too uniformly structured to be common in many scientific applications, with computer vision being a notable exception. Therefore, to better understand performance in practice, we also conduct benchmarks on two real-world datasets: a DNA electrophoresis model and an acoustic vibrobox simulation, both taken from The University of Florida sparse matrix collection \citep{davis:etal:2011}. Figure \ref{fig:real_graphs} shows the adjacency structure of these matrices. We note that while both exhibit some degree of structure, the DNA example presents a much higher degree of regularity and is more similar to a grid graph than the vibrobox example.

\subsection{Trail selection results}
\label{sec:trail_selection_results}
The results of our graph decomposition benchmarks are shown in Table \ref{tab:benchmark_results}. In each example, the medians strategy performs as well or better than the other three methods. Additionally, the naive edge-wise decomposition strategy performs substantially worse overall than any of the trail strategies. For the two graphs with a high degree of structure, namely the grid graph and DNA model, our medians strategy significantly outperforms all of the other methods. When the graphs are less structured, the benefits of clever decomposition fade. We also note that while the random trails perform better on the grid than the pseudo-tour, it is important to remember that the former is an iterative approach that requires much higher preprocessing time, as noted in Section \ref{sec:decomposition}.

\begin{figure}[tbh]
\centering
    \begin{subfigure}{.39\textwidth}
      \centering
      \includegraphics[width=\textwidth]{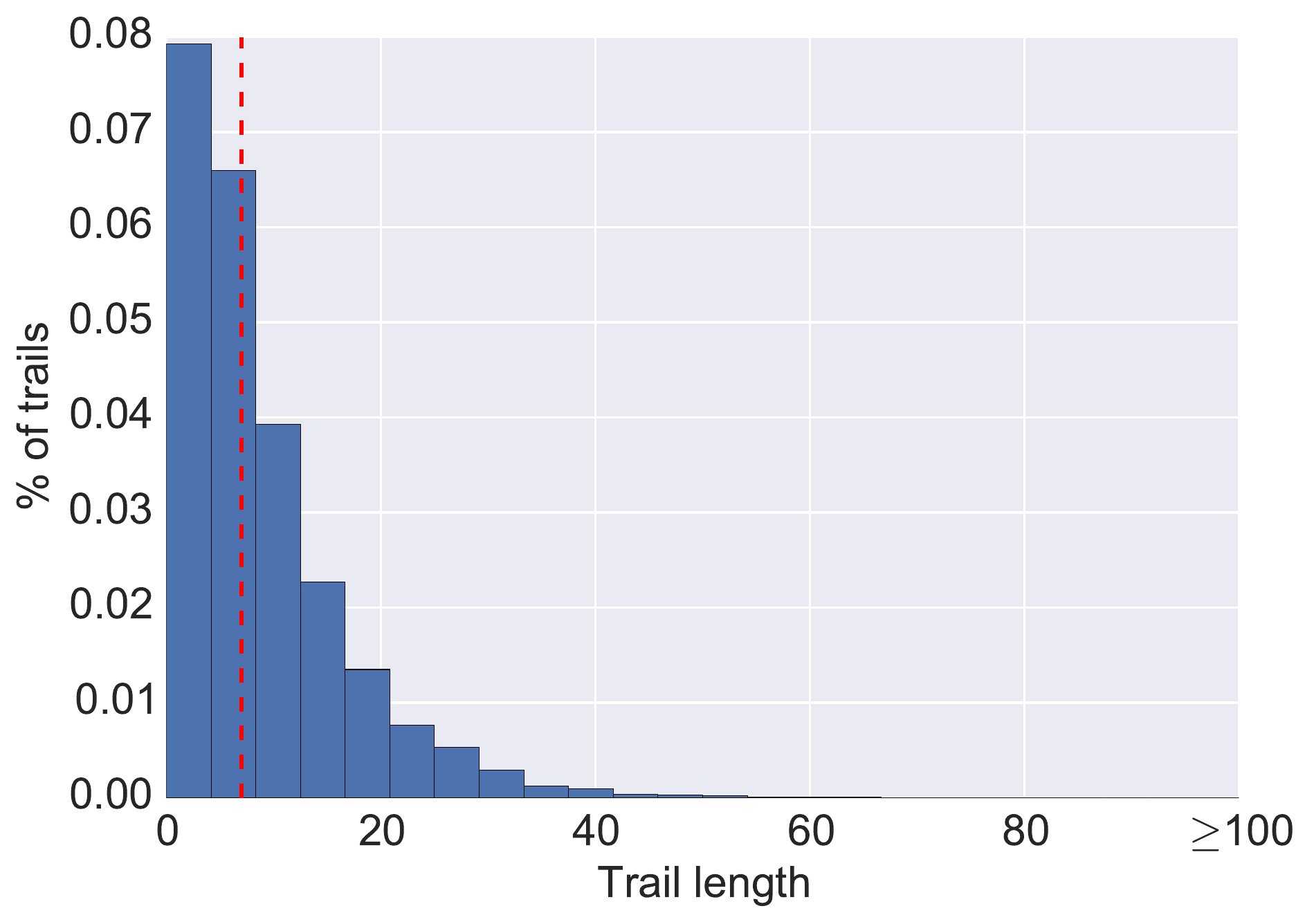}
      \caption{Random graph, pseudo-tour trails}
      \label{fig:trail_distributions_random_tour}
    \end{subfigure}
    \begin{subfigure}{.39\textwidth}
      \centering
      \includegraphics[width=\textwidth]{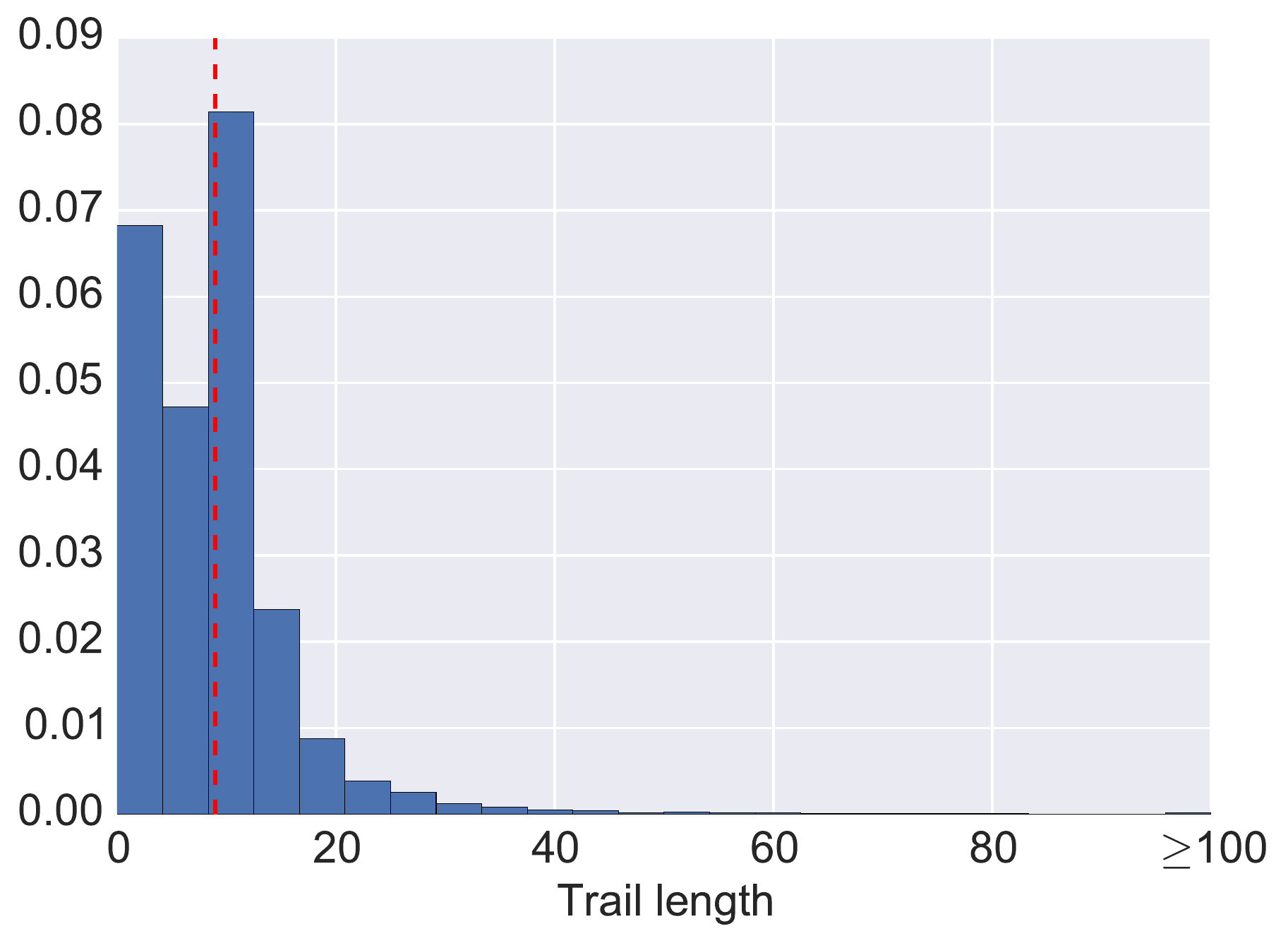}
      \caption{Random graph, median trails}
      \label{fig:trail_distributions_random_medians}
    \end{subfigure}%
    \\
    \begin{subfigure}{.39\textwidth}
      \centering
      \includegraphics[width=\textwidth]{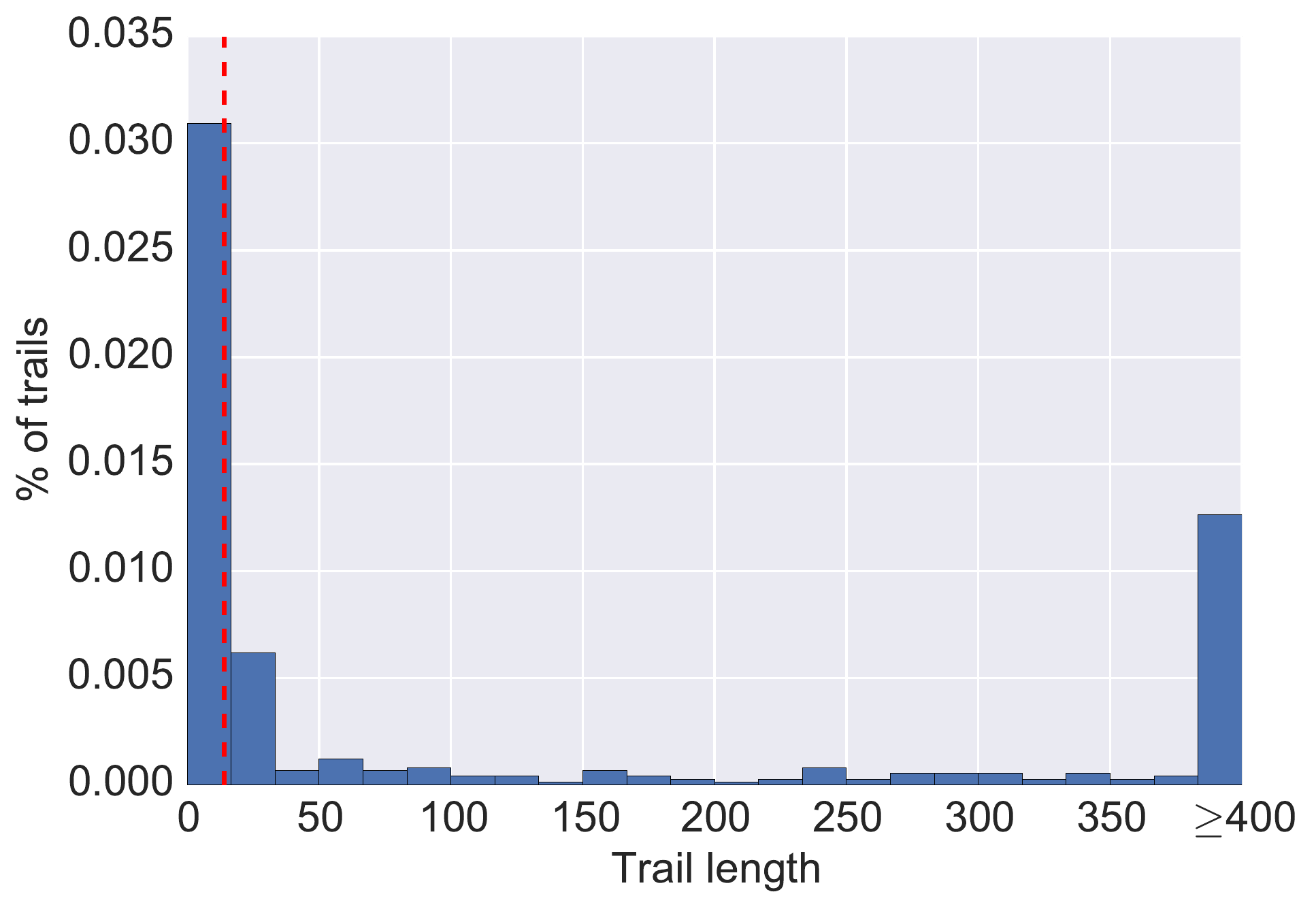}
      \caption{Grid graph, pseudo-tour trails}
      \label{fig:trail_distributions_grid_tour}
    \end{subfigure}
    \begin{subfigure}{.39\textwidth}
      \centering
      \includegraphics[width=\textwidth]{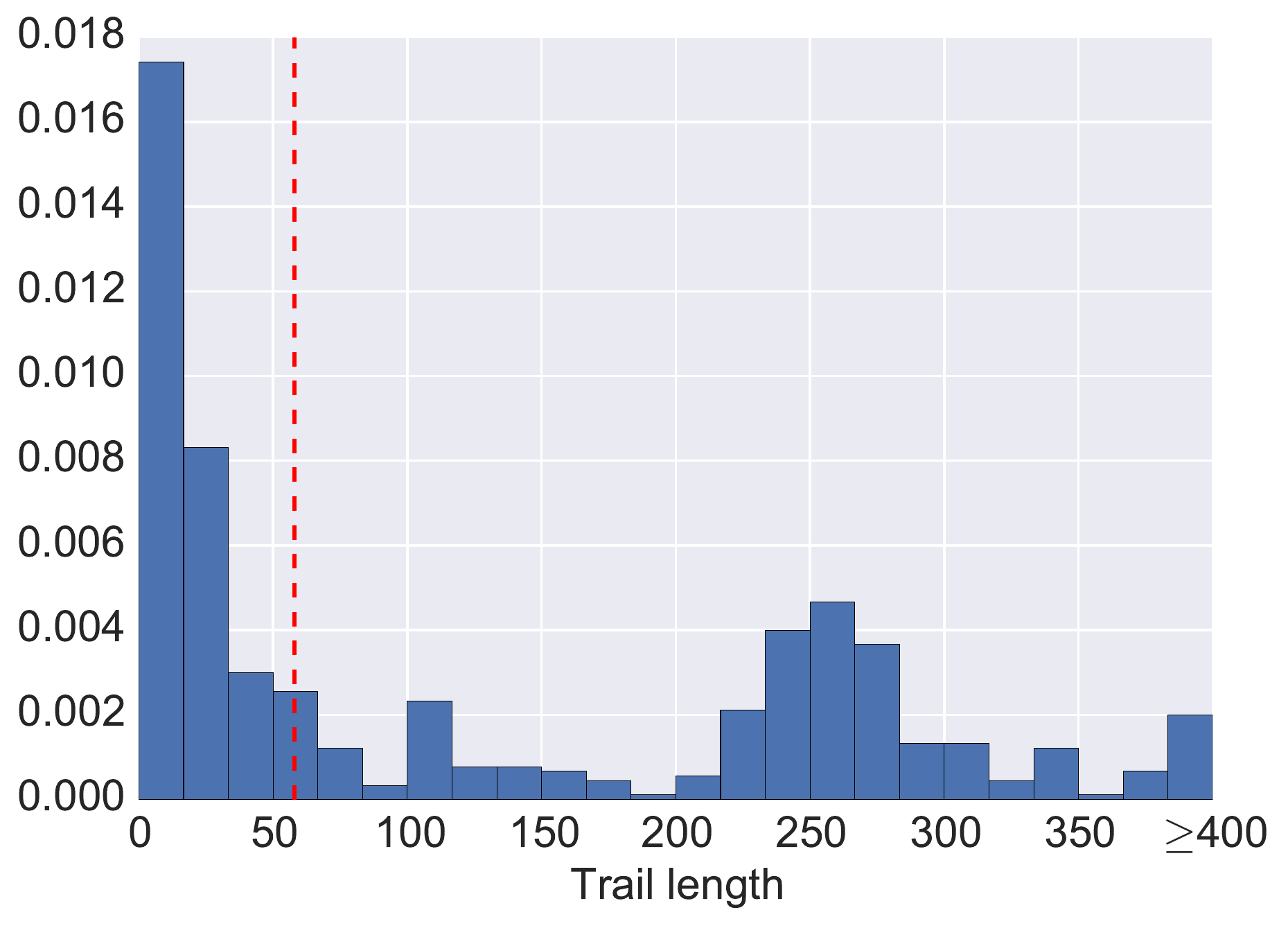}
      \caption{Grid graph, median trails}
      \label{fig:trail_distributions_grid_medians}
    \end{subfigure}%
    \\
    \begin{subfigure}{.39\textwidth}
      \centering
      \includegraphics[width=\textwidth]{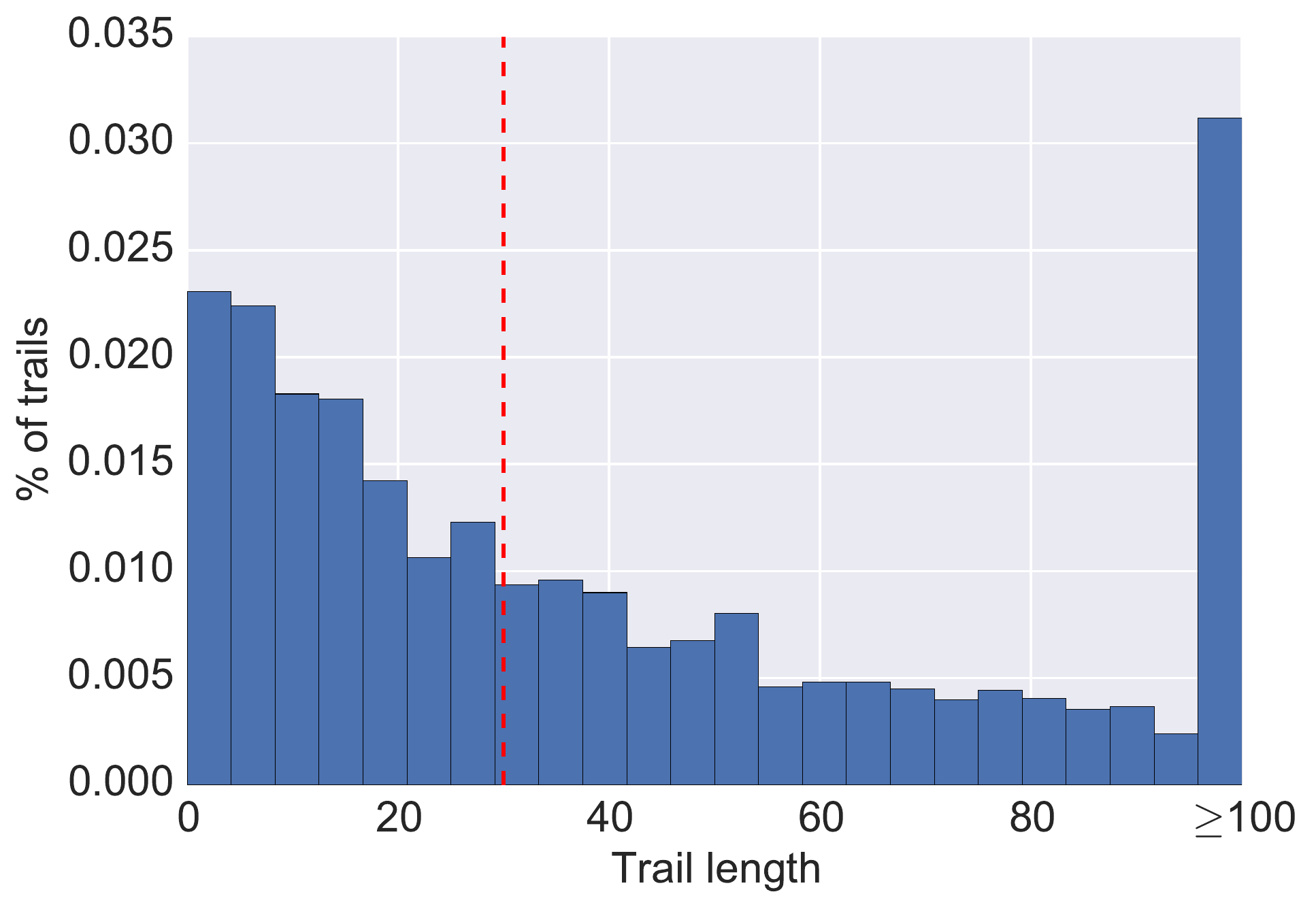}
      \caption{Vibrobox, pseudo-tour trails}
      \label{fig:trail_distributions_vibrobox_tour}
    \end{subfigure}
    \begin{subfigure}{.39\textwidth}
      \centering
      \includegraphics[width=\textwidth]{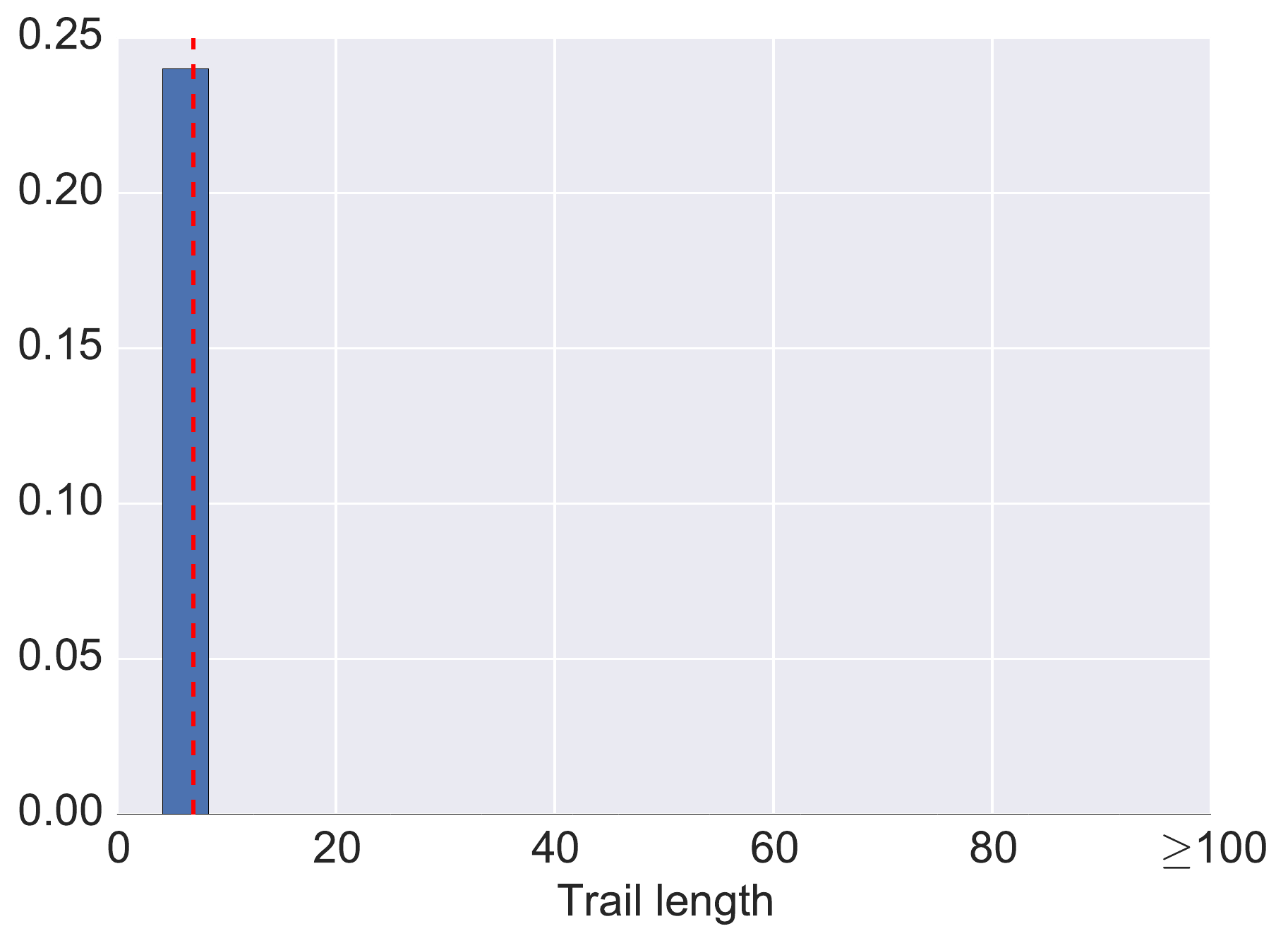}
      \caption{Vibrobox, median trails}
      \label{fig:trail_distributions_vibrobox_medians}
    \end{subfigure}%
    \\
    \begin{subfigure}{.39\textwidth}
      \centering
      \includegraphics[width=\textwidth]{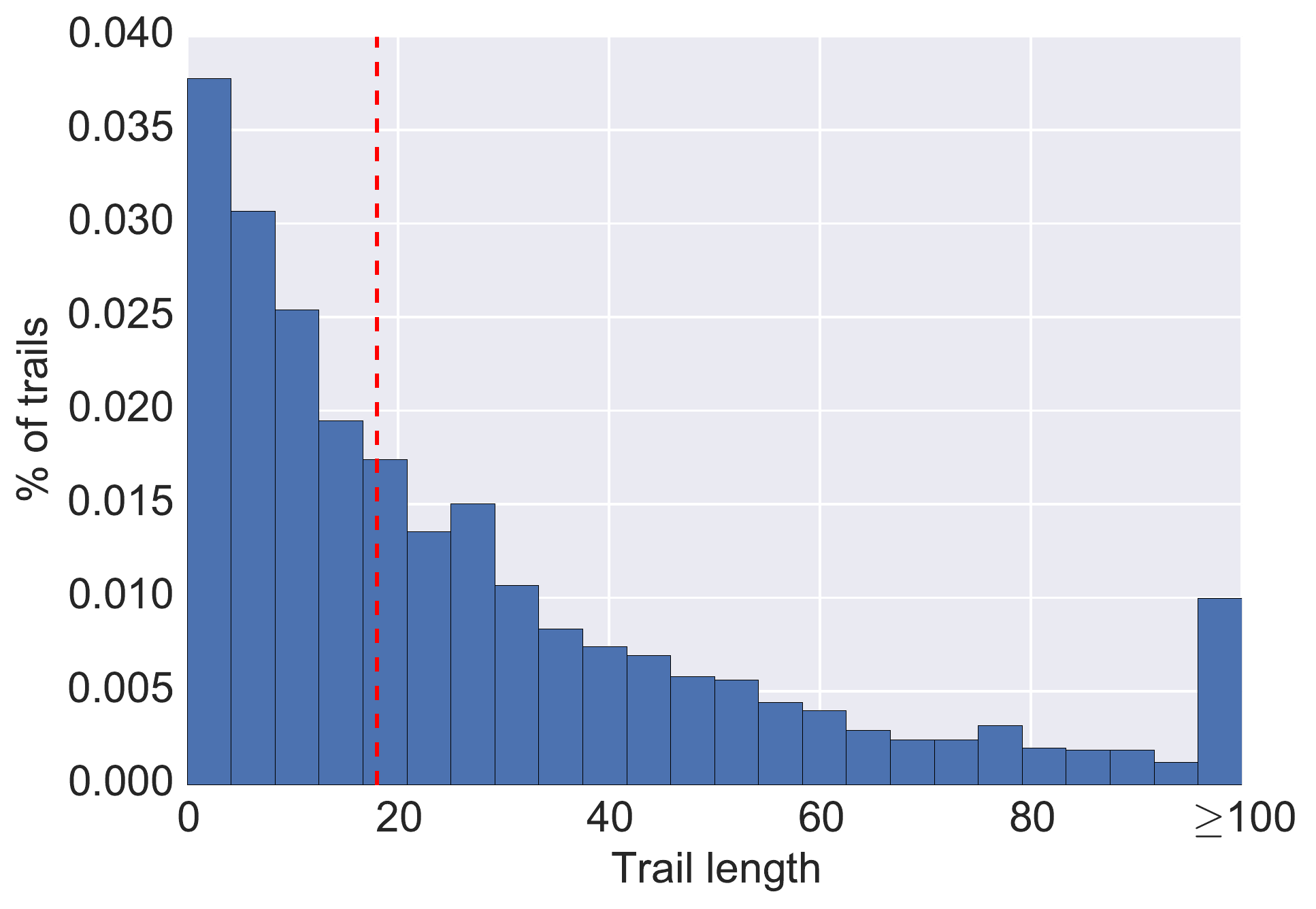}
      \caption{DNA electrophoresis, pseudo-tour trails}
      \label{fig:trail_distributions_dna_tour}
    \end{subfigure}
    \begin{subfigure}{.39\textwidth}
      \centering
      \includegraphics[width=\textwidth]{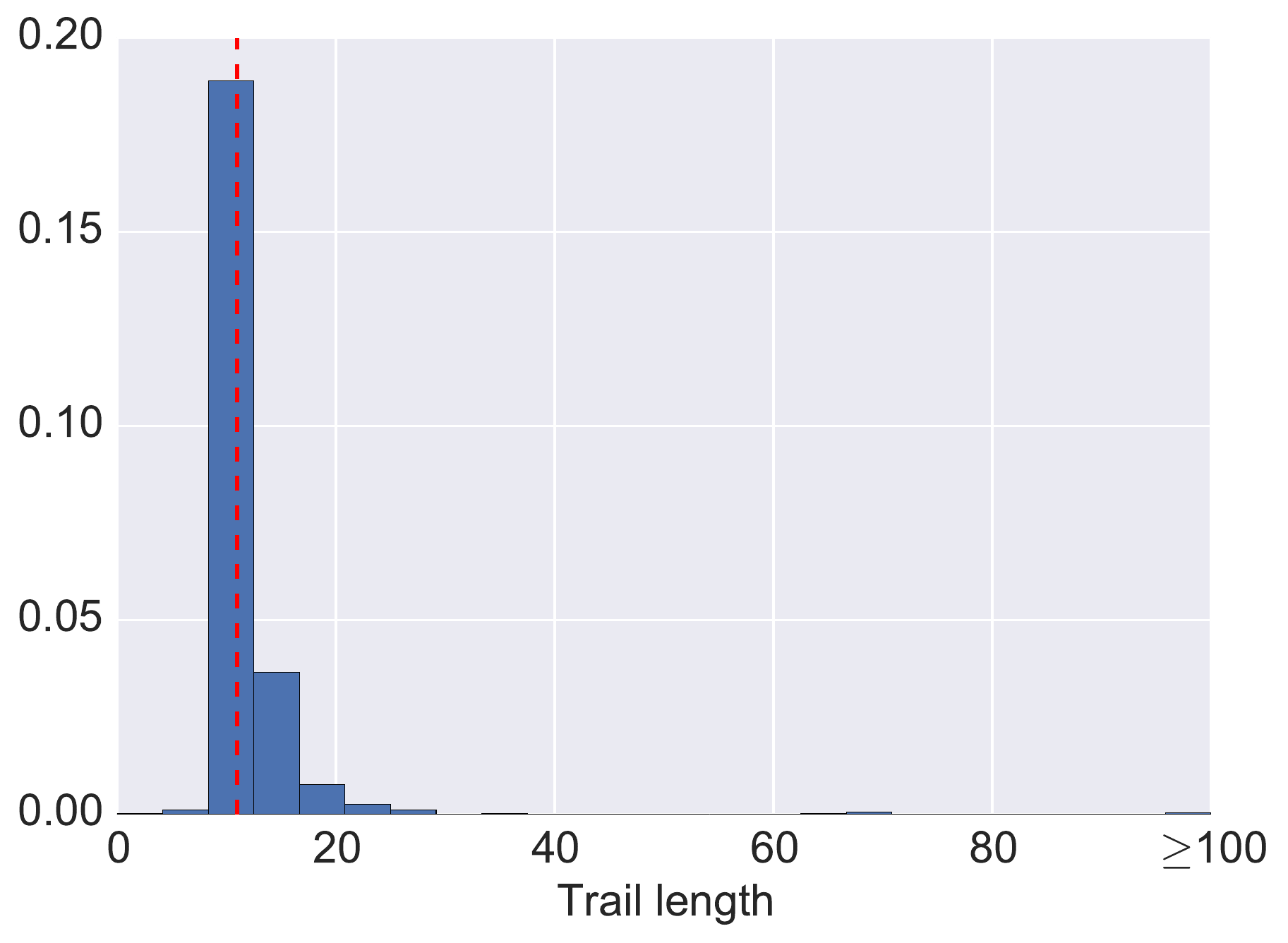}
      \caption{DNA electrophoresis, median trails}
      \label{fig:trail_distributions_dna_medians}
    \end{subfigure}%
\caption{Trail length distributions across the different benchmark graphs; dashed lines are median trail lengths.}
\label{fig:trail_distributions}
\end{figure}

To help shed potential light on the cause of the varying performances, Figure \ref{fig:trail_distributions} shows the distributions of trails for the pseudo-tour and medians strategies across the four benchmark graphs. On random graphs (\ref{fig:trail_distributions_random_tour} and \ref{fig:trail_distributions_random_medians}) the distributions look very similar, with little-to-no balancing aid from the median selection heuristic. When the graph is highly structured, as in the grid graph (\ref{fig:trail_distributions_grid_tour} and \ref{fig:trail_distributions_grid_medians}), the median selection helps to balance out the overall trail set and pushes the overall size of the median trail up. In the vibrobox example (\ref{fig:trail_distributions_vibrobox_tour} and \ref{fig:trail_distributions_vibrobox_medians}) the trails are better balanced by the medians strategy, but at the cost of a much shorter median trail length, likely negating any potential gains from the better balanced trails. Finally, in the DNA example (\ref{fig:trail_distributions_dna_tour} and \ref{fig:trail_distributions_dna_medians}) the medians strategy pays slightly in terms of typical trail length, but the drastic improvement in balance leads to a reasonable performance improvement as noted in Table \ref{tab:benchmark_results}.

\begin{figure}[tbh]
\centering
  \includegraphics[width=0.5\textwidth]{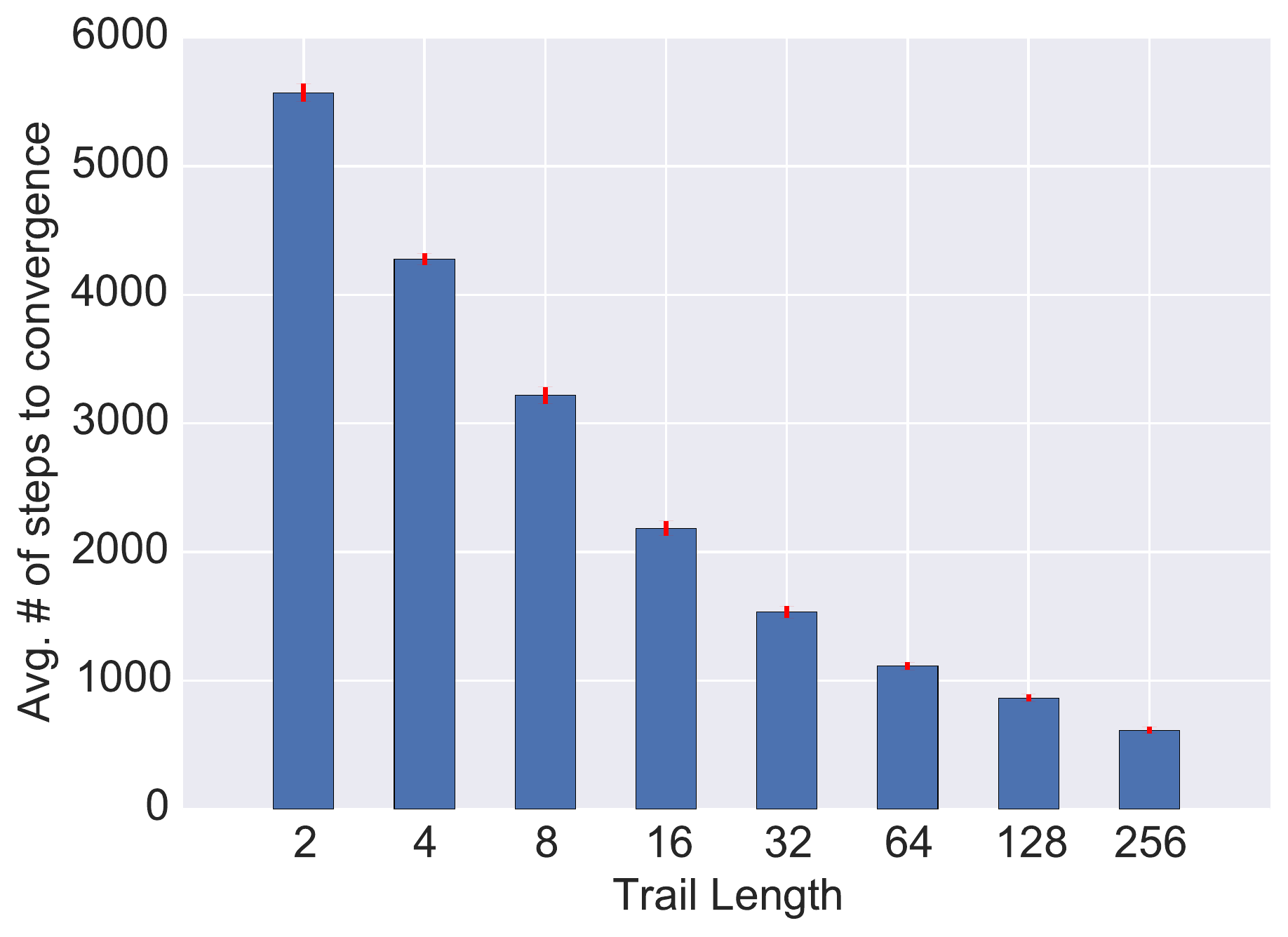}
  \caption{Performance degradation in a $256\times256$ grid graph example as the row and column trails shrink in length.}
  \label{fig:row_col_length}
\end{figure}

This line of reasoning is further supported by Figure \ref{fig:row_col_length}, which shows the performance of splitting the row and column hand-engineered trails on a $256\times256$ grid graph. For this experiment, we recursively break each trail in half and run it to convergence on the same graph. As the figure shows, the length of the trails is directly related to the convergence rate of the algorithm.

\begin{table}
    \centering
    \begin{tabular}{ | l | l | l | l | l | l |}
    \hline
    Graph & Nodes & Pseudo-tour & Medians & Random & Edges \\ \hline
    Random & 50625 & 178 (6) & 186 (6) & 183 (9) & 222 (9) \\
    Grid & 50625 & 11502 (856) & \textbf{6254 (494)} & 9005 (535) & 15228 (723) \\
    Vibrobox & 12328 & 111 (0) & 111 (0) & 111 (1) & 227 (1) \\
    DNA & 39082 & 379 (31) & \textbf{275 (29)} & 359 (31) & 365 (32) \\
    \hline
    \end{tabular}
    \caption{Benchmark results for different trail decomposition strategies. Values reported are the mean number of steps to convergence, with standard errors in parentheses.}
    \label{tab:benchmark_results}
\end{table}

\subsection{Results against other methods}
\label{sec:other_method_results}
\begin{figure}[tbh]
\centering
    \begin{subfigure}{.49\textwidth}
      \centering
      \includegraphics[width=\textwidth]{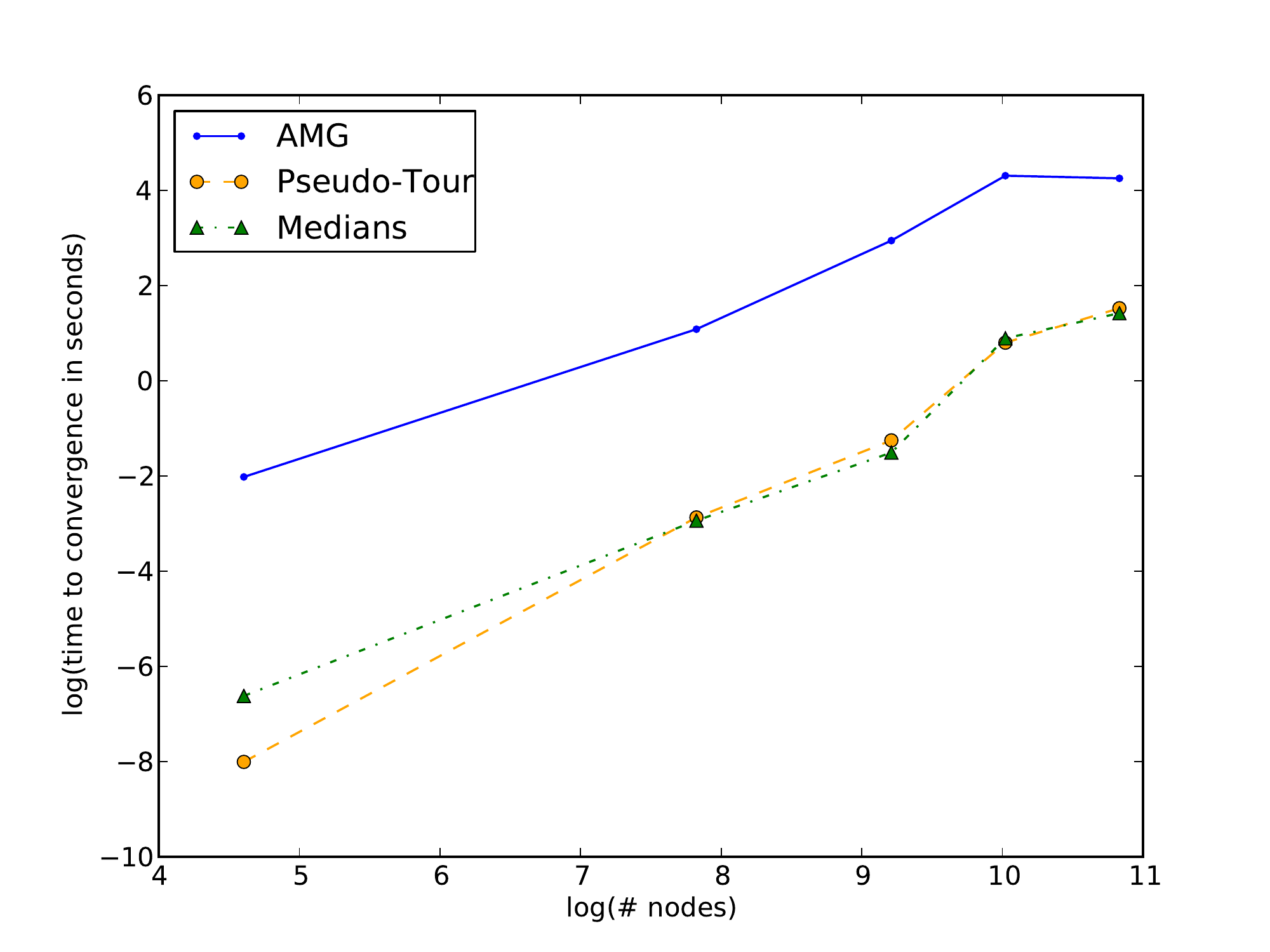}
      \caption{Random graph}
      \label{fig:synthetic_results_random}
    \end{subfigure}
    \begin{subfigure}{.49\textwidth}
      \centering
      \includegraphics[width=\textwidth]{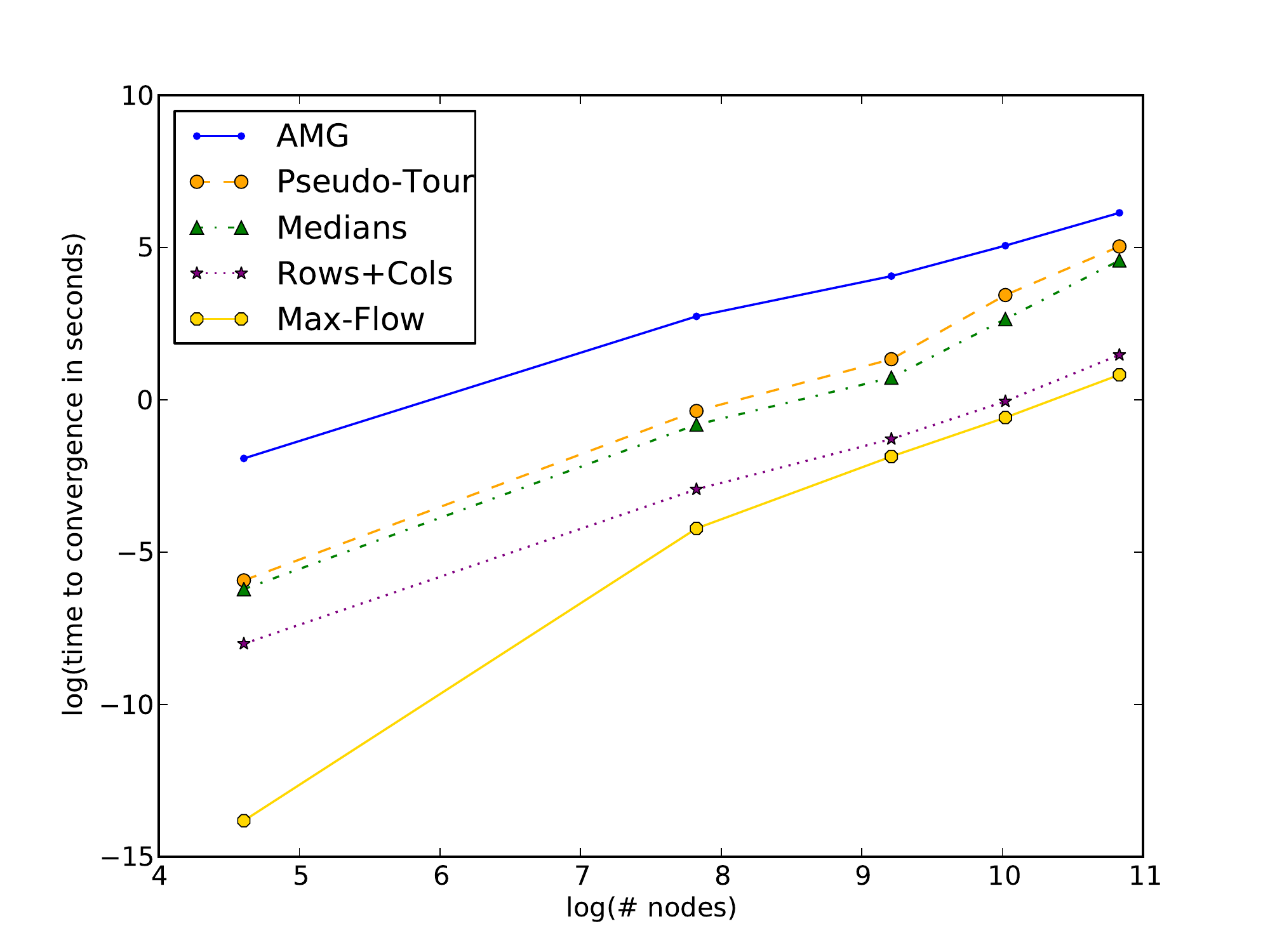}
      \caption{Grid graph}
      \label{fig:synthetic_results_grid}
    \end{subfigure}%
\caption{Log-log plot of comparing performance of the two trail decomposition strategies to the alternative approaches. In both cases, the AMG approach is substantially slower. On random sparse graphs both trails are effectively equal, where as on grid graphs the medians heuristic converges approximately $25\%$ faster.}
\label{fig:synthetic_results}
\end{figure}

Figure \ref{fig:synthetic_results} shows the results of a comparison of the two trails methods versus the alternative implementations on random (\ref{fig:synthetic_results_random}) and grid (\ref{fig:synthetic_results_grid}) graphs. In both cases, we see that the AMG approach takes substantially longer than either trail strategy (note this is a log-log plot). On random graphs (Figure \ref{fig:synthetic_results_random}), there is little difference in performance between the two trail strategies, possibly due to the lack of underlying structural regularity in the graph. On the other hand, on grid graphs the medians strategy converges approximately $25\%$ faster than the pseudo-tour trails. We also note that the simple rows and columns hand-engineered strategy is very competitive with the specialized max-flow method which only works on the limited class of spatial graphs and with the specific squared loss function; our method retains none of these constraints.

%% file: discussion.tex
We have presented a novel approach to solving the fused lasso on a generic graph by the use of a graph decomposition strategy. Experiments demonstrate that our algorithm performs well empirically, even against specialized methods that handle only a limited class of graphs and loss functions. The flexibility and speed of our algorithm make it immediately useful as a drop-in replacement for GFL solvers currently being used in scientific applications (e.g., \citep{tansey:etal:2014}).

There remain two outstanding and related questions regarding our trail decomposition approach: 1) what are the precise properties of a graph decomposition that lead to rapid convergence of our ADMM algorithm, and 2) how can we (efficiently) decompose a graph such that the resulting trails contain these properties? While we have presented some intuitive concepts around the idea of a balanced set of long trails, we are far from answering these questions with certainty. It may be possible to leverage recent work on ADMM convergence rate analysis \citep{nishihara:etal:2015} and in particular work on network models \citep{fougner:etal:2015} to this end. At current, our approaches are purely heuristics driven by empirical success and we plan to explore ways to decompose graphs in a more principled manner in future work.

%% file: spg.tex
An alternative approach to solving \eqref{eqn:gfl_loss_function} is to iteratively solve a smooth approximation \citep{nesterov:2005} of the true non-smooth loss function. This approach has been adapted to similar problems as the GFL, for instance the ``graph-guided fused lasso'' problem was solved via Smoothing Proximal Gradient (SPG) \citep{chen:etal:2012}. Note that SPG is designed to solve problems of the form $f(\mathbf{x}) + g(\mathbf{x}) + h(\mathbf{x})$, where $f(\mathbf{x})$ is a smooth convex function, $g(\mathbf{x})$ is convex, and $h(\mathbf{x})$ is the $L_1$ penalty on $\mathbf{x}$. The $h(\mathbf{x})$ term is not present in \eqref{eqn:gfl_loss_function} and the ``graph-guided fused lasso'' could alternatively be named the ``sparse GFL'', in keeping with the lasso nomenclature (see, for example, the sparse group lasso \citep{friedman:etal:2010}). Nevertheless, if we set the $h(\mathbf{x})$ penalty weight to zero, we recover the following algorithm for solving the GFL:

\begin{algorithm}[htb]
\caption{Gradient smoothing algorithm for the GFL.}
\label{alg:gradient_smoothing_gfl}
\begin{algorithmic}[1]
    \Require Observations $\mathbf{y}$, sparse $|E| \times |V|$ edge differences matrix $D$, error tolerance $\epsilon$
    \Ensure Graph-smoothed $\boldsymbol\beta$ values
    \State $\mu \gets \frac{\epsilon}{|E|}$
    \While{$\boldsymbol\beta$ not converged}
        \State $\boldsymbol\alpha^{t+1} \gets S(\frac{D \boldsymbol\beta^t}{\mu})$
        \State $\grad_{\boldsymbol\beta} (f+\tilde{g}) = \boldsymbol\beta^t - \mathbf{y} + D^T \boldsymbol\alpha^{t+1}$
        \State $\boldsymbol\beta^{t+1} \gets \boldsymbol\beta^t - \eta \grad_{\boldsymbol\beta} (f+\tilde{g})$
    \EndWhile
    \State
    \Return $\boldsymbol\beta$
\end{algorithmic}
\end{algorithm}

In the above algorithm, $S$ is the elementwise truncation operator on the interval $[-1, 1]$, $\tilde{g}$ is the smooth approximation to $g$, and $\eta$ is the typical gradient descent stepsize (determined via backtracking line search in our implementation).

\begin{figure}
\centering
\includegraphics[width=\textwidth]{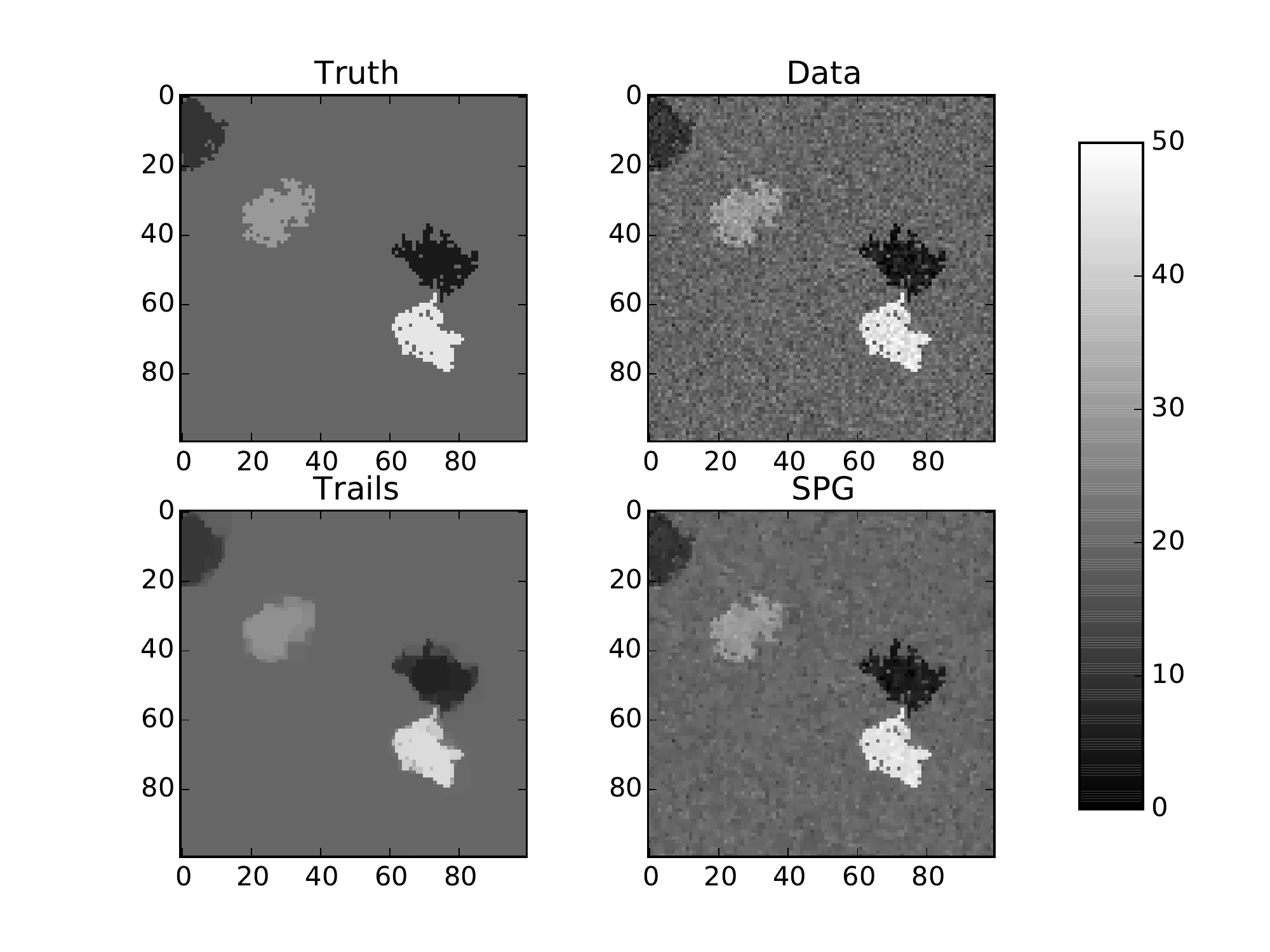}
\caption{Comparison of the different methods for solving the GFL problem. The SPG method recovers a solution that is substantially under-smoothed.}
\label{fig:gradient_smoothing_fail}
\end{figure}

While Algorithm \ref{alg:gradient_smoothing_gfl} converges in the limit \citep{nesterov:2005}, in practice it often gets stuck in a saddle point in our experiments. Figure \ref{fig:gradient_smoothing_fail} shows an example of a $100\times100$ grid-fused lasso problem solved to $10^{-6}$ precision using Algorithm \ref{alg:gradient_smoothing_gfl} and our trail-based approach. The trail method clearly recovers nearly the true underlying parameters, while the SPG-type method is substantially under-smoothed, particularly with respect to the background noise in the image.

